%% file: main.tex
\documentclass[10pt,twocolumn,letterpaper]{article}

\usepackage[pagenumbers]{wacv} 

\input{preamble}

\definecolor{wacvblue}{rgb}{0.21,0.49,0.74}
\usepackage[pagebackref,breaklinks,colorlinks,allcolors=wacvblue]{hyperref}

\def\confName{WACV}
\def\confYear{2026}

\title{Adversarial Pseudo-replay for Exemplar-free Class-incremental Learning}

\author{Hiroto Honda\\
Independent Researcher\\
}

\begin{document}
\maketitle
\input{sec/0_abstract}    
\input{sec/1_intro}
\input{sec/2_relatedwork}
\input{sec/3_method}

\input{sec/4_1_experiment_settings}
\input{sec/4_2_benchmark}
\input{sec/4_3_ablation}
\input{sec/4_4_analysis}

\input{sec/5_conclusion}
\clearpage
{
    \small
    \bibliographystyle{ieeenat_fullname}
    \bibliography{main}
}
\clearpage
\appendix
\input{supp}

\end{document}

%% file: preamble.tex
\usepackage{multirow}
\usepackage{algorithm} 
\usepackage{algpseudocode} 
\usepackage{subcaption}

%% file: sec/0_abstract.tex
\begin{abstract}
Exemplar-free class-incremental learning (EFCIL) aims to retain old knowledge acquired in the previous task while learning new classes, without storing the previous images due to storage constraints or privacy concerns.
In EFCIL, the plasticity-stability dilemma, learning new tasks versus catastrophic forgetting, is a significant challenge, primarily due to the unavailability of images from earlier tasks.  
In this paper, we introduce adversarial pseudo-replay (APR), a method that perturbs the images of the new task with adversarial attack, to synthesize the pseudo-replay images online without storing any replay samples. 
During the new task training, the adversarial attack is conducted on the new task images with augmented old class mean prototypes as targets, and the resulting images are used for knowledge distillation to prevent semantic drift. 
Moreover, we calibrate the covariance matrices to compensate for the semantic drift after each task, by learning a transfer matrix on the pseudo-replay samples.
Our method reconciles stability and plasticity, achieving state-of-the-art on challenging cold-start settings of the standard EFCIL benchmarks.
Code is available at \url{https://github.com/hirotomusiker/APR-EFCIL}.
\end{abstract}

%% file: sec/1_intro.tex
\section{Introduction}
\label{sec:intro}

\begin{figure}
  \centering
  \includegraphics[width=.99\linewidth]{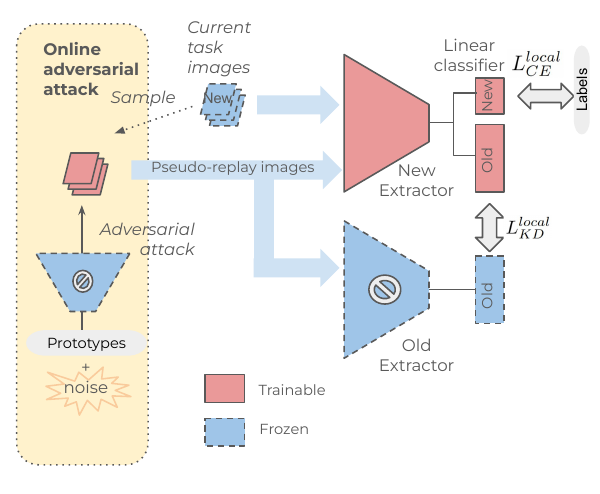}
  \caption{Adversarial Pseudo Replay. The images from the new task are transformed into old-task data via adversarial attack in an \textit{online} manner. Local (logits-based) knowledge distillation using the pseudo-replay images and preserved network prevent the target extractor from semantic drift.}
  \vspace{-1.0em}
  \label{fig:aprfig1}
\end{figure}

Recent rapid advances in deep learning rely on a one-time training using a static dataset. However, in dynamic environments, data often arrive in a non-stationary stream format. The ability to dynamically accumulate the new knowledge is referred to as continual (incremental, life-long) learning \cite{wang2024comprehensive, qu2025recent}. 
Class-incremental learning (CIL) \cite{deepclsurvey} addresses the setting where a group of new class data is only available in the current task and the data from the previously seen classes are not accessible.
The knowledge acquired in the previous tasks is overwritten by the new information, a phenomenon known as catastrophic forgetting \cite{french1999catastrophic}.
Exemplar-free (Non-exemplar) class-incremental learning methods \cite{zhu2021prototype, li2024fcs, goswami2024fecam, lwf, ssre, zhu2021class} tackles the issue by storing old class prototypes --- usually class-wise mean features --- to maintain knowledge without storing raw images (exemplars) in order to avoid privacy and storage issues. 
However, even with the use of prototypes, the plasticity-stability dilemma --- trade-off between acquiring knowledge from new tasks (plasticity) and avoiding catastrophic forgetting (stability) --- remains unresolved.

The major cause of the catastrophic forgetting is the change in the feature extractor, also known as \textit{semantic drift} \cite{yu2020semantic} occurring when learning a new task. This leads to inconsistency between the preserved prototypes and features of past tasks extracted by the updated extractor.
There are two types of approaches for semantic drift: (a) stabilization of the feature extractor and (b) calibrating the prototypes.

LwF \cite{lwf} addresses (a) by introducing knowledge distillation to ensure the network produces similar logits as the preserved old task network on the new task images. FeCAM \cite{goswami2024fecam} also addresses (a) by freezing the feature extractor and employing a network-free Mahalanobis classifier. While the method shows strong stability, it lacks the ability to learn new tasks (plasticity) and performs poorly under the cold-start (or small-start) settings \cite{magistri2024elastic, goswami2024resurrecting} where only limited amount of data are available at the initial task.
We also tackle (a) with knowledge distillation in a more effective way than \cite{lwf}, by transforming new task images into pseudo-replay samples.

Approaches addressing (b) include prototype calibration methods such as \cite{yu2020semantic, li2024fcs, 2024taskrecency}, which transfer prototypes or covariance matrices into the updated feature space. The gap between old and new feature spaces are typically estimated using the new task data and preserved feature extractor. ADC \cite{goswami2024resurrecting} generates synthetic samples through adversarial attack on the new data, for more accurate drift estimation.

In this paper, we address (a) by generating pseudo-replay samples via adversarial attack. 
These samples offer a more effective basis for knowledge distillation than the new task samples alone, in order to prevent extractor's drift from the old task representation space.
Synthetic data can be generated by adversarial attack \cite{goswami2024resurrecting} and stored at the beginning of the new task, or via an external generator network that synthesizes old task data during the new task training \cite{smith2021always}. 
However, both approaches require additional storage space and are thus incompatible with the constraints of exemplar-free CIL. 
To address this, we propose generating pseudo-replay images in an \textit{online} manner, by conducting the adversarial attack during the new task training. 
Before the new task training, we sample the \textit{indices} of new task images and augmentation policies as candidate data, whose features are close to the prototype of each class. 
The recorded augmentation is applied deterministically throughout the new task training.
The online adversarial attack is conducted with old-class target prototypes as targets. 
Inspired by PASS \cite{zhu2021prototype}, we add Gaussian noise to the target prototypes to increase diversity of the resulting pseudo-replay samples. 
Both new-task and pseudo-replay images are passed through the current and old feature extractors, and knowledge distillation \cite{lwf, smith2021always} is performed to minimize the gap of the features. 
As a result, our APR paradigm effectively mitigate the extractor's semantic drift without requiring additional storage space.

We also tackle (b) by calibrating the prototypes and covariance matrices after each task using adversarially generated samples, inspired by \cite{2024taskrecency, goswami2024resurrecting}. A transfer matrix trained for each class using the adversarially generated samples calibrates a covariance matrix via simple matrix multiplication.

Finally, we address an overlooked issue: non-negligible amount of storage required for covariance matrices. We demonstrate that low-rank approximation via singular value decomposition can reduce this overhead without degrading performance.
\\

We summarize our contributions as follows:  
\begin{itemize}
    \item To the best of our knowledge, this is the first work to introduce the adversarial pseudo-replay, which perturbs new-task samples with adversarial attacks to synthesize the pseudo-replay images \textit{online} to mitigate semantic drift, without storing any replay samples.
    \item We further leverage adversarially perturbed samples to calibrate the covariance matrices after each task, by learning transfer matrices to boost Mahalanobis distance-based EFCIL performance.
    \item Our proposed EFCIL pipeline, APR, achieves state-of-the-art results in challenging cold-start settings of CIFAR100, TinyImageNet and ImageNet-Subset, and competitive or best performances on warm-start CIFAR100 and TinyImageNet.
\end{itemize}

%% file: sec/2_relatedwork.tex
\section{Related Work}
\label{sec:related}

To date, a plethora of class-incremental learning methods has been proposed under various data availability scenarios.
In this paper, we focus on exemplar-free class incremental learning (EFCIL) where only class mean features (prototypes) and an old model preserved in the previous task are available during the new task.

\subsection{Semantic Drift Mitigation Methods}
LwF \cite{lwf} leverages knowledge distillation between new-model and old-model logits to mitigate semantic drift of the updated network. PASS \cite{zhu2021prototype} leverages multi-view samples generated by image rotation to learn robust representations in a self-supervised manner. Feature-level knowledge distillation is adopted to prevent semantic drift, and the linear classifier is stabilized with the Gaussian-augmented prototypes. The method is further improved by PASS++ \cite{zhu2024pass++} with hardness-aware prototype augmentation.

Some approaches favor stability and freeze the feature extractor to avoid semantic drift.  SSRE \cite{ssre} proposes self-sustaining expansion and prototype selection strategy. FeTrIL \cite{petit2023fetril} and FeCAM \cite{goswami2024fecam} show excellent performance by leveraging Mahalanobis distance with covariance matrices. These methods do not require incremental training of the feature extractor, however, they face challenges in assimilating new knowledge, especially in cold-start (or small-start) settings \cite{goswami2024resurrecting, magistri2024elastic} where the first task contains only a limited amount of data.
SEED \cite{rypesc2024divide} employs the mixture-of-experts paradigm and selects the optimal expert based on the class distributions.
On the other hand, we allow the feature extractor to be updated during the incremental tasks while maintaining the stability by preventing semantic drift with pseudo-replay.
IL2A \cite{zhu2021class} generates old features using covariance matrices to cope with the mismatch between updated extractor and classifier.

As the feature extractor is updated in the new task, the preserved prototypes become incompatible with the new feature space. The incompatibility is mitigated by computing the feature space gap using the new task samples \cite{yu2020semantic}. ADC \cite{goswami2024resurrecting} further perturbs the new task images by adversarial attacks to calculate the gap more accurately. 
FCS \cite{li2024fcs} introduces a transfer network to calibrate the prototype features for new feature spaces.
AdaGauss \cite{2024taskrecency} addresses dimensional collapse by introducing low-rank projection and anti-collapse loss, and calibrates the covariances with an adapter network after each task to compensate for semantic drift.

\subsection{Generation-based Methods} \cite{smith2021always, gao2022r, qiu2024dual} synthesize old-class samples using model inversion techniques \cite{yin2020dreaming}. DiffClass \cite{meng2025diffclass} employs diffusion models to synthesize previous-task data and fine-tunes them to align their distributions with the sample domains across all tasks. 
However, a powerful generator requires storage and may remember (leak) the sensitive data \cite{gao2022r}. 
In contrast, some methods do not use a generator network and synthesize pseudo samples by applying perturbations to new-task data \cite{goswami2024resurrecting} or combination of the new data and replay exemplars \cite{kumari2022retrospective}.
Our method APR also leverages adversarial attacks on the new task images, but in an online manner during the new task training, to mitigate semantic drift of the feature extractor without storing or fine-tuning exemplars or an external generator network.

%% file: sec/3_method.tex
\section{Method}

\subsection{Adversarial Pseudo Replay : Overview}\label{sec:paradigm}

The training paradigm of our adversarial pseudo-replay is shown in Fig. \ref{fig:aprfig1} and Algorithm \ref{alg:apr}. 
At the initial task $t=0$, the dataset $D_0$ that consists of  the images belonging to the classes $c\in C_0$ is available. The feature extractor $f^0$ and the linear classifier $g^0$ are trained on $D_0$ with cross-entropy loss $L_{CE}$. 
The prototype $\mu_c$ with dimension of $d$ and covariance $\Sigma_c$ whose size is $(d, d)$ are calculated for each class using the trained extractor $f^0$ and $D_0$.

At the subsequent task $t\in{1, 2, 3,...T-1}$, the new task dataset $D_{t}$ for the new class group $C_{t}$ is available and the old data are discarded. In EFCIL setting, the extractor $f^{t-1}$ is preserved along with $\mu_c$ and $\Sigma_c$  $(c\in C_{0:t-1})$. 
During the new task, the new data from $D_t$ are perturbed by adversarial attack (Sec. \ref{sec:candidate}, \ref{sec:adv}) with $\mu_c $ as a target in each iteration, to obtain the pseudo-replay samples of the previous classes $C_{0:t-1}$. 
$f^t$ and $g^t$ are trained with local cross-entropy loss $L_{CE}^{local}$ and local knowledge distillation loss $L_{KD}^{local}$ (Sec. \ref{sec:prkd}) to reconcile new knowledge acquisition (plasticity) and semantic drift mitigation (stability).

\begin{algorithm}[tb]
  \caption{\emph{APR}: Adversarial Pseudo Replay\label{alg:apr}}
\begin{algorithmic}[1]
  \State {\bfseries Initialize:} Training data ($D_0, D_1, \dots, D_{T-1}$), feature extractor $f^0$, linear classifier $g^0$
  \State Train $f^0$ and $g^0$ on $D_0$ with $L_{CE}$
 \State Calculate prototype $\mu_c$ and covariance $\Sigma_c$ for {$c \in C_{t}$}  
  \For {$t=1,2,3,\dots T-1$}
      \For {$c \in (C_0,...C_{t-1})$}
          \State Sample $I_{c} \subseteq \{1, \dots, |D_t|\}$ with $f^{t-1}$ and $\mu_c$ 
          \State Record policy $P_{c}$ \Comment {(Sec. \ref{sec:candidate})}
      \EndFor
      \State Form $D_{APR}$ with $D_{t}$ and $\{I_c, P_c\}_{c\in (C_0,...C_{t-1})}$ 
      \For{epoch $e = 1$ to $E$}
          \For{batch $B_{t} \subset D$ and $B_{APR} \subset D_{APR}$}
              \State Perturb $B_{APR} \rightarrow B^{\dagger}_{APR}$ \Comment {APR (Sec. \ref{sec:adv})}
              \State Train $f^t$ and $g^t$ on $B_{t}, B^{\dagger}_{APR}$ with eq. \ref{eq:loss}
          \EndFor
      \EndFor
      \For {$c\in (C_0,...C_{t-1})$}
          \State Obtain $D^{\dagger}_{c}$ with ADC 
          \State Train $W$ with $f^{t-1}$, $f^{t}$ and $D^{\dagger}_{c}$ \Comment {(Sec. \ref{sec:covcal})}
          \State Calibrate $\mu_c$ and $\Sigma_c$ \Comment {(eq. \ref{eq:covcal})} 
      \EndFor 
      \State Calculate $\mu_c$ and $\Sigma_c$ for {$c \in C_{t}$}  
  \EndFor
\end{algorithmic}
\end{algorithm}

After training on $D_t$, the prototypes $\mu_c$ and covariance matrices $\Sigma_c$ of the old classes $c\in (C_0,...C_{t-1})$ are calibrated to compensate for semantic drift of the extractor. The calibration is conducted with the adversarially perturbed images to accurately measure the difference between the features extracted by $f_t$ and $f_{t-1}$. The transfer matrix $W$ is trained as a single layer perceptron so that the old covariance matrices are calibrated to the new feature space with a simple matrix multiplication (Sec. \ref{sec:covcal}).

At test time, three types of classifiers are available; i) linear classifier $g^t$, ii) nearest class mean (NCM) classifier using $\mu_c$ and iii) Mahalanobis classifier using $\mu_c$ and $\Sigma_c$ .

\subsection{Candidate Sampling}\label{sec:candidate}

At the beginning of the task ${t}$, the new task data $D_t$, old network $f^{t-1}$, prototypes $\mu_c^{t-1}$ and $\Sigma_c^{t-1}$ are available. 
We wish to synthesize the old task data with adversarial perturbation \cite{goswami2024resurrecting}, without relying on an external generator network \cite{meng2025diffclass}, and use the generated pseudo-replay data for knowledge distillation \cite{smith2021always}.
One option to achieve the scheme is to prepare pseudo-replay images before the main training loop of the task $T$ , and keep them throughout the task to be loaded along with the new task dataset $D_{t}$.
However, since one of the premises of EFCIL is limited storage space, keeping all the pseudo-replay images (e.g. 224 $\times$ 224 $\times$ 3 for ImageNet-Subset) during the task is not acceptable.

Therefore, in APR we only keep \textit{indices} and \textit{augmentation parameters} of the candidate samples from the new task dataset $D_{t}$ before the task.
More specifically, for every old classes {$c \in (C_0,...C_{t-1})$}, we load and apply augmentations on the images from $D_{t}$, extract the features with the old extractor $f^{t-1}$ and calculate the Euclidean distance between the features and the prototype $\mu_c$:

\begin{equation}
    \label{eq:distance}
    d_i = \| f^{t-1}(P_i(x_i)) - \mu_c \|_2   , x_i \subseteq D_t,
\end{equation}
where $P_i$ is the augmentation policy with the parameters that are randomly set for each sample. The parameters include random crop coordinates, horizontal flip flag and auto-augment \cite{cubuk2018autoaugment} policy parameters.
The typical number of the policy parameters is less than 30, which we consider satisfies the EFCIL's storage limitation demand. 
More details can be found in the Supplementary Material.

Then we pick $k$ indices and augmentation policy parameters of the samples from the smallest distance:
\begin{equation}
    \vspace{-0.4em}
    \label{eq:argsort}
    I_c = \operatorname{argsort}({d})_{[:k]}.
\end{equation}
We gather $I_c$, $P_c=\{ P_i\}_{i\in I_c}$ and the pseudo-labels $Y_c=\{c\}_{i\in I_c}$ for the all old classes {$c \in (C_0,...C_{t-1})$} and form a pseudo-replay candidate dataset $D_{APR}$ that reproduces the candidate images without storing actual image data.

\subsection{Adversarial Attack}\label{sec:adv}

In each training iteration, a batch of  data $B_{t} \subset D_t$ for task-$t$ and another batch of pseudo-replay candidate data $B_{APR} \subset D_{APR}$ are loaded and augmented. Both $B_t$ and $B_{APR}$ consist of the new task images but the latter is loaded using the candidate indices $I_{c}$ and deterministic augmentation policy $P_{c}$.

Subsequently, the images $x\in B_{APR}$ are perturbed into $x_{adv}\in B^{\dagger}_{APR}$ with adversarial attack.
Following \cite{goswami2024resurrecting}, we move the feature of $x$ towards the corresponding prototype $\mu_c$. For each $x\in B_{APR}$, the feature is extracted with the frozen old task network $f_{t-1}$. The loss function $L$ calculates the distance between the feature and the prototype $\mu_c$, which is back-propagated through $f_{t-1}$ and the input image $x$ is updated: 

\begin{equation}
    \label{eq:attack}
    x_{adv}\leftarrow x-\alpha\frac{\nabla_x L(f_{t-1}(x), \mu_c)}{\| \nabla_x L(f_{t-1}(x), \mu_c)\| _2^2}, 
\end{equation}
where $\alpha$ is the attack magnitude. The attack is repeated $N_{attack}$ times.
We do not ensure the similarity between $x_{adv}$ and $x$ \cite{goodfellow2014explaining} or clip pixel value after the attack \cite{goswami2024resurrecting}.
To encourage generalization of training, Gaussian noise $r\mathcal{N}(0, 1)$ is applied to $\mu _c$ for every perturbation target. Similar to prototype augmentation \cite{zhu2021prototype} , the magnitude $r$ is calculated from trace of covariances:
\begin{equation}
    \label{eq:noisemag}
    r=\sqrt{\sum_c{\frac{\text{Tr}(\Sigma_c^{t-1})}{d}}},
\end{equation}
where $\Sigma_c^{t-1}$ is the covariance matrix calibrated before task $t$ (Sec. \ref{sec:covcal}) and $d$ is the feature dimension.
The resulting images $x_{adv}\in B^{\dagger}_{APR}$ are directly used for the following knowledge distillation without postprocessing.

\subsection{Knowledge Distillation with Pseudo-Replay}\label{sec:prkd}

We adopt the knowledge distillation paradigm proposed in \cite{lwf}, that imposes the local cross-entropy loss $L_{CE}^{local}$ and local knowledge distillation loss $L_{KD}^{local}$ on the batches of $B_t$ and $B^{\dagger}_{APR}$ .
\begin{equation}
    \label{eq:localce}
    L_{CE}^{local}=\text{CE}(g^t_n(f^t(x)), y^\dagger), x\in B_t,
\end{equation}
\begin{equation}
    \label{eq:localkd}
    L_{KD}^{local}=\text{KD} (g^{t}_o(f^t(x)), g^{t-1}_o(f^{t-1}(x))), x\in \{B_t ,B^{\dagger}_{APR}\},
\end{equation}
where $\text{CE}$ is the cross-entropy loss, $\text{KD}$ is the knowledge distillation loss \cite{lwf}, $g_n$ and $g_o$ are the split linear classifier corresponding to new-task and old-task classes, and $y^\dagger$ is the relative class index in the current task classes.
\\
In summary:
\begin{itemize}
    \item $L_{CE}^{local}$ : applied on the new-class logits, using only new-task data,
    \item $L_{KD}^{local}$ : applied on the old-class logits, using new-task data and pseudo-replay data.
\end{itemize}
The learning target at task $t(>0)$ is summarized below:
\begin{equation}
    \label{eq:loss}
    \mathcal{L} = \mathcal{L}_{CE}^{local} + \lambda_{KD}^{local} \mathcal{L}_{KD},
\end{equation}
where $\lambda_{kd}$ is the loss weight parameter.

\subsection{Prototype and Covariance Calibration}\label{sec:covcal}
 
The prototypes calculated in the previous task is calibrated to mitigate the effect of semantic drift caused by the update of $f^t$.  
We leverage perturbed samples $D^{\dagger}_c$ corresponding to each class for calibration. 
To transfer the covariance matrix to the updated representation space, the $(d, d)$ transfer matrix $W$ is trained for each class. 
More specifically, the samples $x^{\dagger}_c$ from the dataset $D^{\dagger}_c$ are fed to the old and new feature extractor $f^{t-1}$ and $f^{t}$, and $W$ is trained so that the gap between $f^t(x^{\dagger}_c)$ and $Wf^{t-1}(x^{\dagger}_c)$ is minimized. 
The covariance matrix $\Sigma_c$ is calibrated by the following: 
\begin{equation}
    \label{eq:covcal}
    \mu_c^t = \mu_c^{t-1}+\Delta_c, \ \ \Sigma_c^t = W\Sigma_c^{t-1}W^T,
\end{equation}
where $\Delta_c$ stands for mean feature differences between $f^t(x^{\dagger}_c)$ and $f^{t-1}(x^{\dagger}_c)$. This covariance calibration is simpler and faster than the existing work \cite{2024taskrecency} which applies the transform on the data sampled from the multivariate Gaussian ($\mathcal{N}(\mu_c^{t-1}, \Sigma_c^{t-1})$) to calculate $\Sigma_c^{t}$.

At test time, Mahalanobis distance between the test data feature and the multivariate distribution $\mathcal{N}(\mu_c^t, \Sigma_c^t)$ is calculated. 
Following \cite{goswami2024fecam}, we shrink and normalize the covariance matrix before distance calculation:

\begin{equation}
    \label{eq:shrink}
    \vspace{-0.3em}
    \Sigma_s = \Sigma+\gamma_1 V_1I + \gamma_2 V_2I ,
\end{equation}
\begin{equation}
    \label{eq:normalize}
    \vspace{-0.3em}
    \Sigma^{*}(i, j) = \frac{\Sigma_s(i, j)}{\sqrt{\Sigma_s(i, i)\Sigma_s(j, j)}},
\end{equation}
where $V_1$ and $V_2$ are the average on-diagonal and off-diagonal variance of $\Sigma$ respectively. 

\subsection{Covariance Decomposition}\label{sec:decomposition}

The covariance matrices have rich information from the past classes but require more storage than prototypes. The size of a covariance matrix $(d, d)$ corresponds to $d$ prototypes. \cite{2024taskrecency} suggests that the feature representation incurs dimensional collapse throughout the tasks. The sparsity of the features indicates a possibility to compress the covariances into storage-efficient data. We leverage singular value decomposition (SVD) for obtaining the rank-$k$ ($<d$) representation; 
\begin{equation}
    \label{eq:svd}
    \Sigma_k = U_kS_kV_k .
\end{equation}
The decomposed matrices $U_k$, $S_k$, $V_k$ have the total size of $2kd + k^2$, which is considerably smaller than the full covariance matrix.

%% file: sec/4_1_experiment_settings.tex
\section{Experiment}

\begin{figure*}
  \centering
  \includegraphics[width=1.00\linewidth]{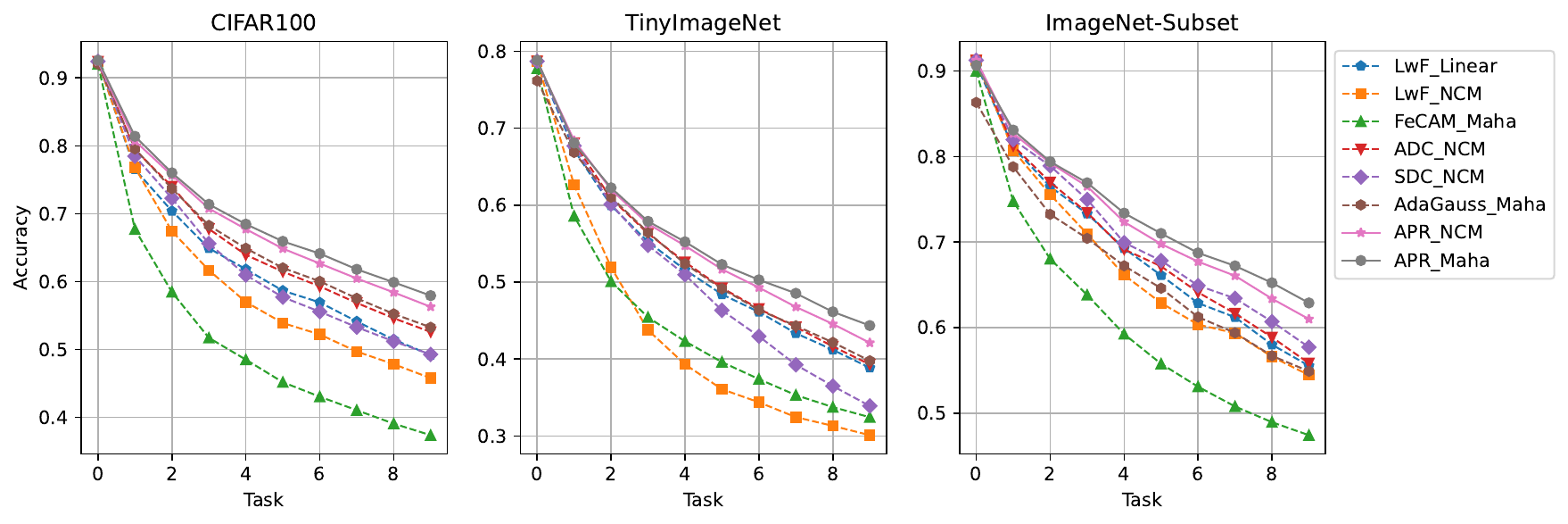}
  \vspace{-2.0em}
  \caption{Accuracy transition across all tasks on the cold-start settings.  All the results are averaged over three random seeds using our implementation. Best viewed in color.}
  \label{fig:transition}
\end{figure*}

\begin{table*}[t]
\centering
\resizebox{0.94\textwidth}{!}{
\begin{tabular}{@{\kern0.5em}llcccccccccccc@{\kern0.5em}}
        \toprule
        \multirow{3}{*}
            && \multicolumn{4}{c}{CIFAR-100}
            && \multicolumn{4}{c}{TinyImageNet}
            && \multicolumn{2}{c}{ImageNet-Subset}
        \\ \cmidrule(lr){3-6} \cmidrule(l){7-11} \cmidrule(l){12-14}
            && \multicolumn{2}{c}{$T$=5}
            & \multicolumn{2}{c}{$T$=10}
            &
            & \multicolumn{2}{c}{$T$=5}
            & \multicolumn{2}{c}{$T$=10}
            &
            & \multicolumn{2}{c}{$T$=10}
            
        \\ \midrule
Method& Classifier& $A_{inc}$ & $A_{last}$ & $A_{inc}$ & $A_{last}$ && $A_{inc}$ & $A_{last}$ & $A_{inc}$ & $A_{last}$ && $A_{inc}$ & $A_{last}$ 
        \\ \midrule
SDC \cite{yu2020semantic} & NCM & 64.82 & 54.94 & 58.02 & 41.36  && 50.82 & 40.05 & 40.46 & 27.15 && 65.83 & 43.72 \\
ADC \cite{goswami2024resurrecting} & NCM & 69.62 & 59.14 & 61.35 & 46.48 && 50.94 & 41.00 & 43.04 & 32.32 && 67.07 & 47.58 \\
AdaGauss \cite{2024taskrecency} & Maha & - & - & 60.20 & 46.10 && - & - & 50.60 & 36.50 && 65.00 & 51.10 \\
\hline
\hline
Joint & Linear & 83.25 & 78.69 & 84.18 & 78.69 && 69.84 & 65.71 & 70.54 & 65.71 && 87.48 & 85.39  \\
\hline
LwF \cite{lwf} & Linear & 71.20 & 59.39 & 63.66 & 49.19 && 59.67 & 48.66 & 53.06 & 38.87 && 69.47 & 55.53  \\
LwF \cite{lwf} & NCM & 67.89 & 53.96 & 60.48 & 45.79 && 50.30 & 35.26 & 44.07 & 30.10 && 67.83 & 54.44  \\
FeCAM \cite{goswami2024fecam} & Maha & 62.90 & 48.08 & 52.61 & 37.82 && 53.51 & 40.95 & 46.01 & 33.48 && 61.02 & 47.83 \\
SDC \cite{yu2020semantic} & NCM & 71.97 & 60.18 & 63.68 & 49.28 && 58.86 & 46.85 & 51.12 & 33.90 && 71.16 & 57.71  \\
ADC \cite{goswami2024resurrecting} & NCM & 72.53 & 61.41 & 66.24 & 52.59 && 59.00 & 47.69 & 53.75 & 39.30 && 69.98 & 55.83  \\
AdaGauss \cite{2024taskrecency} & Maha & 72.95 & 61.74 & 66.66 & 53.27 && 58.52 & 47.20 & 53.53 & 39.68  && 67.96 & 55.07   \\
\hline
APR & Linear  & 72.58 & 61.05 & 67.20 & 53.18 &&   59.06 & 47.05 & 54.70 & 41.78  && 70.93 & 58.81  \\
APR & NCM & \textit{\color{blue}74.39} & \textit{\color{blue}64.39} & \textit{\color{blue}69.00} & \textit{\color{blue}56.29} && \textit{\color{blue}60.14} & \textit{\color{blue}49.55} & \textit{\color{blue}55.58} & \textit{\color{blue}42.11} && \textit{\color{blue}73.00} &  \textit{\color{blue}60.99} \\
APR & Maha & \textbf{\color{red}74.99} & \textbf{\color{red}65.55} & \textbf{\color{red}69.96} & \textbf{\color{red}57.94} && \textbf{\color{red}60.74} & \textbf{\color{red}50.77} & \textbf{\color{red}56.39} & \textbf{\color{red}44.35}  &&  \textbf{\color{red}73.86} & \textbf{\color{red}62.88}  \\
\hline
\end{tabular}
}
\vspace{-2mm}
\caption{
    Average and final incremental accuracy in \textit{cold-start} EFCIL settings across three benchmarks. Only $1/T$ of the total classes are in the initial task.Experiments below the double lines are the averaged results over three runs with our implementation described in Sec. \ref{subsec:setup}. 'Maha' stands for Mahalanobis classifier. The results of \cite{yu2020semantic} are from \cite{goswami2024resurrecting}. For ImageNet-Subset, the first convolution and MaxPool strides are 1 and 2 following \cite{goswami2024fecam}. 
    \textbf{\color{red}Best} in bold red, \textit{\color{blue}second best} in italic blue.
    }
    \vspace{-4.3mm}
\vspace{2mm}
\label{tab:coldstart_benchmark}
\end{table*}

\begin{table}[ht]
\centering
\begin{subtable}{0.44\textwidth}
    \resizebox{0.99\textwidth}{!}{
    \centering
        \begin{tabular}{@{\kern0.5em}llcccc@{\kern0.5em}}
        \toprule
         & & \multicolumn{2}{c}{$T=6$}& \multicolumn{2}{c}{$T=11$}\\
         \hline
        Method& Classifier& $A_{inc}$ & $A_{last}$& $A_{inc}$ & $A_{last}$\\
        \hline
        PASS \cite{zhu2021prototype} & Linear & 63.84 & 55.67 &  59.87 & 49.03   \\
        PASS$++$ \cite{zhu2024pass++} & Linear & 69.12 & 59.87 & 66.50 & 57.69 \\
        DCMI \cite{qiu2024dual} & Linear  & 67.90 & -  & 66.80 & - \\
        SEED \cite{rypesc2024divide} & Linear & 70.9 & -  & 69.3 & - \\
        \hline
        \hline
        Joint & Linear & 81.43 & 78.69 & 81.43 & 78.69 \\
        \hline
        LwF \cite{lwf} & Linear & 69.74 & 59.32 & 65.40 & 53.10\\
        LwF \cite{lwf} & NCM & 71.82 & 61.96 & 67.72 & 54.75 \\
        FeCAM \cite{goswami2024fecam} & Maha & 71.62 & 62.39 & \textit{\color{blue}71.52} & \textit{\color{blue}62.39} \\
        SDC \cite{yu2020semantic} & NCM & 72.77 & 64.26 & 69.56 & 58.43 \\
        ADC \cite{goswami2024resurrecting} & NCM & 72.74 & 64.36 & 70.32 & 60.27\\
        AdaGauss \cite{2024taskrecency} & Maha & 72.38 & 64.60 & 70.04 & 59.76 \\
        \hline
        APR & Linear & 71.83 & 62.90 & 67.73 & 56.64  \\
        APR & NCM & \textit{\color{blue}73.92} & \textit{\color{blue}66.14} & 71.49 & 62.26 \\
        APR & Maha & \textbf{\color{red}74.48} & \textbf{\color{red}67.30}  & \textbf{\color{red}72.57} & \textbf{\color{red}63.74}  \\ 
        \hline
        \end{tabular}
    }
    \caption{CIFAR100.}
\end{subtable}
\hfill
\begin{subtable}{0.44\textwidth}
    \centering
    \resizebox{0.99\textwidth}{!}{
    \begin{tabular}{@{\kern0.5em}llcccc@{\kern0.5em}}
        \toprule
         & & \multicolumn{2}{c}{$T=6$}& \multicolumn{2}{c}{$T=11$}\\
         \hline
        Method& Classifier& $A_{inc}$ & $A_{last}$ & $A_{inc}$ & $A_{last}$ \\
        \hline
        PASS \cite{zhu2021prototype} & Linear & 49.53 & 41.58 & 47.15 & 39.28 \\
        PASS$++$ \cite{zhu2024pass++} & Linear & 54.13 & 46.93 & 53.14 & 46.66 \\
        DCMI \cite{qiu2024dual} & Linear  & 54.8 & - & 53.9 & - \\
        \hline
        \hline
        Joint & Linear & 67.92 & 65.71 & 67.91 & 65.89 \\
        \hline
        LwF \cite{lwf} & Linear & 60.61 & 52.09 & 55.73 & 42.02  \\
        LwF \cite{lwf} & NCM & 53.10 & 39.35 & 45.31 & 27.70  \\
        FeCAM \cite{goswami2024fecam} & Maha & \textit{\color{blue}60.77} & \textbf{\color{red}53.49} & \textbf{\color{red}60.59} & \textbf{\color{red}53.49} \\
        SDC \cite{yu2020semantic} & NCM & 58.93 & 49.91 & 54.35 & 39.29 \\
        ADC \cite{goswami2024resurrecting} & NCM & 58.91 & 50.54 & 55.18 & 44.38  \\
        AdaGauss \cite{2024taskrecency} & Maha & 58.38 & 50.33 & 56.08 & 46.46 \\
        \hline
        APR & Linear & \textbf{\color{red}60.90} & 52.98 & 57.32 & 45.97 \\
        APR & NCM & 59.81 & 52.06 & 56.67 & 47.30 \\
        APR & Maha & 60.66 & \textit{\color{blue}53.24} & \textit{\color{blue}57.69} & \textit{\color{blue}48.56} \\ 
        \hline
    \end{tabular}
    }
    \caption{Tiny-ImageNet.}
\end{subtable}

\caption{Average and final incremental accuracy in \textit{warm-start} EFCIL in (a) CIFAR100 and (b) Tiny-ImageNet. \textit{Half} of the classes are in the initial task. Experiments below the double lines are the averaged results over three runs with our implementation. Results of \cite{zhu2021prototype} are from \cite{zhu2024pass++}.
\textbf{\color{red}Best} in bold red, \textit{\color{blue}second best} in italic blue.}
\label{tab:warmstart_benchmark}
\end{table}

\subsection{Benchmark Setup}\label{subsec:setup}

\noindent \textbf{Datasets and CIL Settings.}
The benchmarks are conducted on the following three datasets:
\noindent CIFAR100 \cite{krizhevsky2009learning} contains 100 classes, each with 500 (train) and 100 (test) images of (32, 32) resolution. TinyImageNet \cite{le2015tiny} consists of 200 classes each of which has 500 train and 50 test images of (64, 64) resolution. ImageNet-subset contains 100 classes \cite{wang2022foster} selected from ImageNet-1k \cite{deng2009imagenet}, each of which has approx. 1300 train and 50 test images of size (224, 224).
In typical $T$-task EFCIL, the model is trained on half of the total classes at the initial task, and $1/(T-1)$ of the rest are incremented at the following $T-1$ tasks. This setting is referred to as \textbf{warm-start setting}. In contrast, the more challenging and practical \textbf{cold-start setting} was introduced in \cite{goswami2024resurrecting}, where only $1/T$ of the classes are available at the initial task. The stability-favoring methods such as \cite{petit2023fetril, goswami2024fecam} perform well under warm-start settings but struggle in cold-start scenarios due to the limited knowledge acquired at the initial task. We aim to build a method that performs well in both settings.

\noindent\textbf{Evaluation Metric.}
We employ \textit{average incremental accuracy} ($A_{inc}$) and \textit{final accuracy} ($A_{last}$) for evaluation. 
$A_{inc}$ and $A_{last}$ are standard metrics \cite{rebuffi2017icarl, goswami2024fecam, goswami2024resurrecting} to evaluate overall accuracy trends and final performance.
The accuracy up to task $k$ ($A_k$) and $A_{inc}$ are defined as:

\begin{equation}
    \label{eq:fm_msr}
    A_k = \frac{1}{k} \sum_{j=1}^{k} a_{k,j}\ ,\ \ A_{inc} = \frac{1}{K} \sum_{j=1}^{K} A_k,
    \vspace{-0.2cm}
\end{equation}
where $a_{k,j}$ stands for accuracy of class $j$ at task $k$.

\subsection{Implementation Details}\label{sec:impl}

For fair comparison, we fix the following components that affect the training outcome significantly:

\noindent\textbf{\textit{Network}}: Following EFCIL literature \cite{goswami2024fecam,goswami2024resurrecting}, we employ ResNet18 \cite{he2016deep} for feature extractor and split cosine classifier. For ImageNet-Subset, we employ the first convolution with stride=1 and MaxPool down-sampling with stride=2 \cite{goswami2024fecam}, which deals with larger feature maps than the 2-stride convolution counterpart. 
The output feature size is set to 512-dim, except for 64-dim in AdaGauss following \cite{2024taskrecency}.

\noindent\textbf{\textit{Augmentation}}: For CIFAR100, TinyImageNet and ImageNet-Subset, we apply random crop that outputs (32, 32), (32, 32), and (224, 224) images respectively. We do not use a four-view self-supervision setting \cite{zhu2021prototype} or class augmentation \cite{zhu2021class}. 
In CIFAR100, AutoAugment \cite{cubuk2018autoaugment} policies for CIFAR10 are applied after horizontal flipping.

\noindent\textbf{\textit{Optimization}}: 
In the initial task, the model is trained with SGD, learning rate (lr) = 0.1, and weight decay (wd) = 5e-4 for 200 epochs; in incremental stages with lr = 0.01, wd = 2e-4 for 100 epochs (60 for ImageNet-Subset) both under cosine decay. For warm-start, the incremental lr is 0.001.

\noindent\textbf{\textit{CE and KD Loss}}: We apply local CE loss (eq. \ref{eq:localce}) to the new-class part and local KD loss (eq. \ref{eq:localkd}) to the old-class part of the logits following \cite{lwf, goswami2024resurrecting}. The loss weights are set $\lambda_{kd} = 10$ as in \cite{goswami2024resurrecting, lwf} for all the settings.  
\\

\noindent The settings regarding our method are as follows:

\noindent\textbf{\textit{APR}}: The batch size for $B_{APR}$ is set to 32, 64 and 64 for CIFAR100, Tiny-ImageNet and ImageNet-Subset respectively. 
200 new task images and augmentation policies per class are sampled for pseudo-replay. For CIFAR100 and ImageNet-Subset, Autoaugment policies for CIFAR10 and ImageNet are applied after horizontal flip. The attack magnitude $\alpha$ is fixed to 64 and number of iteration $N_{attack}$ is set to 4, 6 and 2 for CIFAR100, Tiny-ImageNet and ImageNet-Subset respectively.

\noindent\textbf{\textit{Covariance Calibration}}: After the task, ADC \cite{goswami2024resurrecting} is performed for each class to generate perturbed samples $D^\dagger$. We train a transfer matrix $W$ as a $(d, d)$ linear layer with learning rate of 1e-4 for 64 epochs.

\noindent\textbf{\textit{Covariance Shrinkage}}: We apply covariance shrinkage (eq. \ref{eq:shrink}) before evaluation. To avoid overfitting to the test data, the shrinkage parameters $\gamma_1$ and $\gamma_2$ in eq. \ref{eq:shrink} are determined by splitting the validation dataset ($N$=50 per class) from the train dataset. 

All the experiments are carried out with a single NVIDIA RTX4070 GPU. Other implementation details are provided in Supplementary Material.

\begin{table}
\resizebox{0.99\linewidth}{!}{
        \begin{tabular}{ccc|ccccl}
            \hline
              Calib- & Pseudo &  Adv.  & \multicolumn{2}{c}{NCM}& \multicolumn{2}{c}{Mahalanobis} &Train\\ 
              ration & Replay &  attack  & $A_{inc}$ & $A_{last}$ & $A_{inc}$ & $A_{last}$ &time\\
            \hline
            \checkmark &           &             & 71.48 & 57.88 & 72.30&  60.47 &0.48\\
            \checkmark &\checkmark &             & 71.07 & 56.59 & 72.03& 58.85 &0.66\\
                       &\checkmark &  \checkmark & 70.00& 56.92& 69.81 & 56.45 &0.93\\
            \checkmark &\checkmark &  \checkmark & \textbf{73.00} &  \textbf{60.99} & \textbf{73.86}& \textbf{62.88} &1.00\\
            \hline
        \end{tabular}
}
    \caption{Ablation study for adversarial pseudo replay and calibration on ImageNet-Subset $T=10$. First convolution and MaxPool strides are 1 and 2 \cite{goswami2024fecam}.}
    \vspace{-1.5em}
    \label{tab:ablation_apr}
\end{table}

\begin{table}
\resizebox{0.85\linewidth}{!}{
    \begin{tabular}{cc|cccc}
        \hline
         Determ. & Prototype & \multicolumn{2}{c}{NCM}& \multicolumn{2}{c}{Mahalanobis}\\ 
         aug. & noise& $A_{inc}$ & $A_{last}$ & $A_{inc}$ & $A_{last}$\\
         \hline
                    & \checkmark & 72.91& 61.46& 73.65&  63.29\\
         \checkmark &            & 72.44 & 59.81 & 73.35& 61.75\\
         \checkmark & \checkmark & \textbf{73.00} &  \textbf{60.99} & \textbf{73.86}& \textbf{62.88}\\
         \hline
    \end{tabular}
    }
    \caption{Ablation study for deterministic augmentation and target prototype noise on ImageNet-Subset $T=10$ setting. First convolution and MaxPool strides are 1 and 2 \cite{goswami2024fecam}.}
    \label{tab:ablation_adv}
\end{table}

%% file: sec/4_2_benchmark.tex
\subsection{Benchmark Results}\label{sec:bench}

First, we report performance comparison on the cold-start EFCIL setting in Table \ref{tab:coldstart_benchmark} and Fig. \ref{fig:transition}. The results of the existing methods above the double horizontal lines are from their papers, and the rest including Joint and APR are evaluated with the averages of experiments across three random seeds. 
The \textit{Joint} stands for the upper-bound accuracy results, where all the old task classes are available at each task. 
APR surpasses all the methods with large margins in $A_{inc}$, +3.3\% or CIFAR-100, +2.9\% for Tiny-ImageNet and +2.7\% for ImageNet-Subset under $T=10$ setting. The improvement of $A_{last}$ is also considerable, +4.7\% in CIFAR100, +4.5\% in Tiny-ImageNet and +5.2\% in ImageNet-Subset ($T=10$), compared with the best value among existing methods. These results validate APR's ability to mitigate semantic drift while adapting to new tasks flexibly. 
FeCAM \cite{goswami2024fecam} and AdaGauss \cite{2024taskrecency} store covariance matrix for each old class, with the dimension of 512 and 64 respectively. Our APR surpasses the methods even with NCM classifier without covariances (+11.8\% over FeCAM and +5.7\% over AdaGauss in $A_{inc}$ of ImageNet-Subset).

Second, Table \ref{tab:warmstart_benchmark} shows the warm-start setting benchmarks, where half of the classes are available at the initial task. 
APR is competitive or best on most settings: 
In CIFAR100, APR with Mahalanobis classifier shows the state-of-the-art performance in both $T=6$ and $T=11$ settings, even surpassing FeCAM that freezes the feature extractor during the incremental stages. 
In Tiny-ImageNet, APR is competitive or best on most settings. FeCAM is slightly better on $A_{inc}$ / $A_{last}$ ($T=11$) and $A_{last}$ ($T=6$) than APR (Maha). However, unlike FeCAM APR does not discard the capability of learning the new task.

APR requires more training time (31.0 hours on Imagenet-Subset $T=10$), compared with FeCAM (2.9h), ADC (13.5h), and AdaGauss (20.1h), due to online adversarial attack and pseudo-replay (see Supp. Material for details). Computational complexity for inference does not increase from ADC or FeCAM, since APR does not modify network architecture or attach additional components.

\begin{table}
    \resizebox{0.35\linewidth}{!}{
    \begin{tabular}{cc|cc}
        \hline
         $\alpha$ & $A_{inc}$ & $A_{last}$ \\ 
        \hline
        8 & 68.88 & 55.35  \\ 
        16 & 69.52 & 57.04  \\ 
        32 & 70.00 & 57.40  \\ 
        \textbf{64} &  \textbf{69.96} & \textbf{57.94} \\ 
        \hline
    \end{tabular}
    }
    \quad
    \resizebox{0.55\linewidth}{!}{
    \begin{tabular}{c|ccc}
        \hline
        loops & $A_{inc}$ & $A_{last}$ & Time (h)\\ 
        \hline
        1 & 69.56 & 56.74 & 3.9 \\
        2 & 70.01 & 57.51 & 4.3 \\
        \textbf{4} & \textbf{69.96} & \textbf{57.94} & \textbf{5.0} \\ 
        6 & 69.89 & 58.05 & 5.8 \\
        \hline
    \end{tabular}
    }
    \vspace{-0.5em}
    \caption{Sensitivity sweep on CIFAR100 ($T=10$). Parameters used in Tab. \ref{tab:coldstart_benchmark} are in bold. ‘Time’ denotes total benchmark runtime.}
    \label{tab:sweep}
\end{table}

\begin{table}
    \begin{center}
    \resizebox{0.65\linewidth}{!}{
        \begin{tabular}{ccc|cl}
            \hline
            SVD $k$& Size& $\gamma_1, \gamma_2$& $A_{inc}$&$A_{last}$\\
            \hline
            N/A& 100$\%$ & 56& 73.37&61.75\\
            64& 26.6$\%$ & 56& 73.36&61.73\\
            8& 3.1$\%$ & 96& 73.23&61.53\\
            \hline
            NCM& - & -& 72.50 &59.91\\
            \hline
        \end{tabular}
        }
    \end{center}
    \vspace{-1.5em}
    \caption{Effect of covariance decomposition on Mahalanobis-based performances on ImageNet-Subset $T=10$. First convolution and MaxPool strides are 1 and 2 \cite{goswami2024fecam}.}
    \label{tab:svd}
\end{table}

\begin{table}[h!]
\centering
    \resizebox{0.79\linewidth}{!}{
    \begin{tabular}{c|cccc}
        \hline
        Component & FeCAM & AdaGauss & APR & APR$_{\text{SVD} k=8}$ \\
        \hline
        Prototypes & 0.18 & 0.02 & 0.18 & 0.18 \\
        Covariances  & 94.32 & 1.44 & 94.32 & 2.97  \\
        Candidate inds  & 0.00 & 0.00 & 0.14 & 0.14 \\
        Aug. Params & 0.00 & 0.00 & 1.08 & 1.08 \\
        \hline
        Sum          & 94.50 & 1.46 & 95.73 & 4.38  \\
        \hline
    \end{tabular}
}
\caption{Comparison of memory/storage usage (MB) at $t=9$ on ImageNet-Subset $T=10$, assuming 90 old classes, 200 candidates per class and 13k new-task images.}
\label{tab:storage}
\end{table}

%% file: sec/4_3_ablation.tex
\subsection{Ablation Study}\label{sec:ablation}

\noindent \textbf{Adversarial Pseudo Replay.} Table \ref{tab:ablation_apr} compares pseudo-replay configurations on ImageNet-Subset cold-start $T=10$ setting. All metrics reflect averages over three random seeds and train time is normalized by the full setting.
Applying pseudo-replay without adversarial attack (row 2) increases training time but contributes to the notable accuracy gain  (+1.9\% $A_{inc}$ in NCM and +1.8\%  in Mahalanobis), highlighting its essential role. 
Ablating pseudo replay entirely (row 1) shows slightly better performance than the second row (0.4\% $A_{inc}$ in NCM and 0.3\%  in Mahalanobis), showing that pseudo replay without adversarial refinement can be counterproductive. Thus, adversarial attack is the key driver of performance.
Table \ref{tab:sweep} confirms that attack effects stabilize around the chosen perturbation magnitude $\alpha$ and repetition $N_{attack}$.

\noindent \textbf{Prototype and covariance calibration.} 
The third row in Table \ref{tab:ablation_apr} omits prototype and covariance calibration after each task. Without this component,  NCM and Mahalanobis classifiers experience significant performance degradation (-3.0\% and -4.0\%, respectively). In this setting prototypes and covariances remain fixed after they are initially created. Although APR mitigates semantic drift of the extractor, it is still crucial to align the prototypes and covariance matrices to the evolving feature space.

\noindent \textbf{Deterministic augmentation and prototype noise.} Table \ref{tab:ablation_adv} investigates the impact of components for generating pseudo-replay samples. Deterministic augmentation applies the same transforms used for candidate selection; without it (top row), random transforms weaken pseudo-replay, causing a small drop in $A_{inc}$ (0.20\%).
Prototype noise \cite{zhu2021prototype}, ablated in the second row, diversifies the pseudo-replay samples and ensures the adversarial attack generalization, similar to data augmentation. 
Removing it reduces both NCM and Mahalanobis accuracy significantly (0.56\% $A_{inc}$ for NCM and 0.51\% for Mahalanobis), highlighting its importance for APR.

%% file: sec/4_4_analysis.tex
\subsection{Analysis}

\begin{figure}
  \centering
  \includegraphics[width=.8\linewidth]{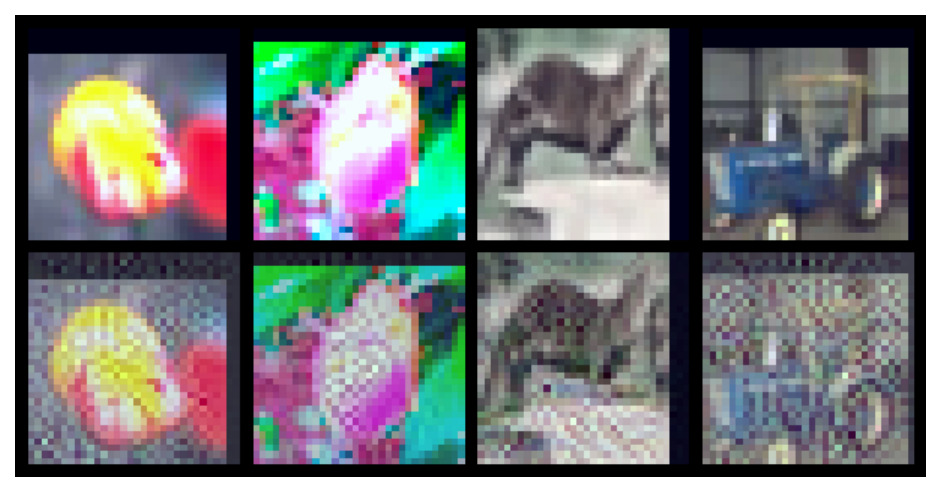}
  \vspace{-1.0em}
  \caption{Example of pseudo-replay samples (top: before attack, bottom: after attack) from CIFAR100 \cite{krizhevsky2009learning}.}
  \label{fig:images}
\end{figure}

\begin{figure}
  \centering
  \includegraphics[width=.99\linewidth]{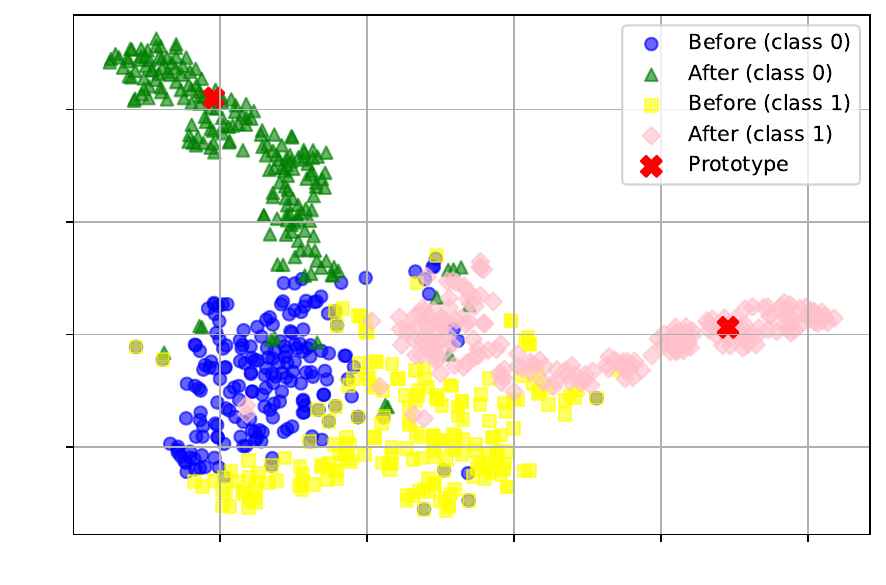}
  \vspace{-1.0em}
  \caption{t-SNE analysis of online adversarial attacks during $t=1$ training on CIFAR100. The features extracted by $f^{t-1}$ and prototypes of two classes are shown. Best viewed in color.}
  \vspace{-1.0em}
  \label{fig:tsne}
\end{figure}

\begin{figure}
  \vspace{-2.0em}
  \centering
  \includegraphics[width=.99\linewidth]{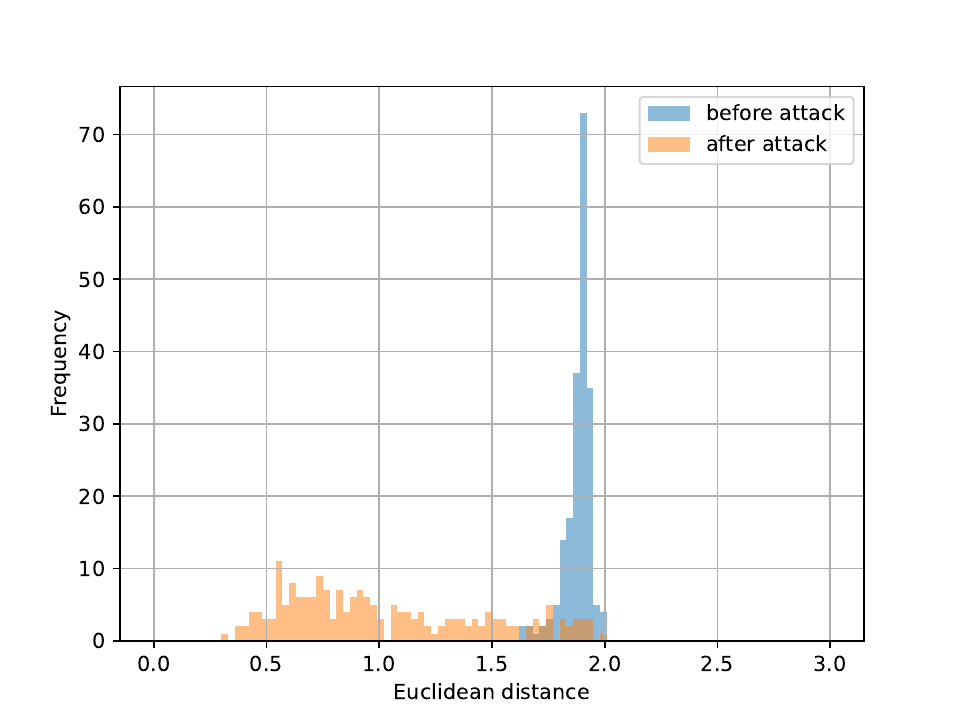}
  \vspace{-1.0em}
  \caption{Euclidean distance distributions between target prototype and image features extracted by $f^{t-1}$ before and after attack. Best viewed in color.}
  \vspace{-1.0em}
  \label{fig:dist}
\end{figure}

\noindent \textbf{Online adversarial attack.}\label{sec:analysis}  The effectiveness of adversarial pseudo-replay samples stems from their proximity to the class prototypes in the feature space.
Fig. \ref{fig:images} illustrates how online adversarial attacks alter the appearance of input images. The perturbations manifest as visible noise patterns, which guide the extracted features toward the target prototypes.
Fig. \ref{fig:tsne} visualizes feature distributions via t-SNE, and Fig. \ref{fig:dist} shows the Euclidean distances to the target prototype before and after the attack.
The results confirm that the features shift significantly closer to their respective prototypes after adversarial attacks, indicating the new task images are effectively transformed into pseudo-replay representations. Additional visualizations are in the Supp. Material.

\noindent \textbf{Covariance Decomposition.} 
Table \ref{tab:svd} shows the effect of covariance decomposition (Sec. \ref{sec:decomposition}). The Mahalanobis accuracy does not deteriorate even at $k=8$ corresponding to size reduction of 97\%. As $k$ gets smaller, more covariance shrinkage is necessary to maintain performance. However this result significantly mitigates the often-overlooked storage demand for covariance matrices. 

\noindent \textbf{Storage Comparison.}
Required storage for each component at the late stage is summarized in Table \ref{tab:storage}. Covariance matrices require the largest space but mitigated with small dimension (AdaGauss) or decomposition (APR$_{\text{SVD} k=8}$). The candidate indices and augmentation parameters are small, enabling storage-efficient pseudo-replay without storing the actual image data (e.g. $>$10GB). See Supp. Material for more details.

%% file: sec/5_conclusion.tex
\section{Conclusion}

In this paper we presented Adversarial Pseudo Replay (APR), a method designed to mitigate semantic drift problem of exemplar-free class-incremental learning. 
Pseudo-replay data are synthesized online during the new-task training using an adversarial attack, and are used for local knowledge distillation to mitigate semantic drift of the feature extractor.
The effectiveness of APR is further enhanced by augmenting target prototypes during adversarial attack.
Mahalanobis classifier shows the best performance with after-task calibration of the covariance matrices, using transfer matrices trained with perturbed samples. 
Our APR significantly improves the plasticity-stability trade-off in both cold-start and warm-start settings, without the need to store replay samples or an external generative network.
\linebreak

\noindent\textbf{Limitations.} APR is storage-efficient, but incurs increased training time due to the use of pseudo-replay data and adversarial attacks. Improving efficiency of adversarial attack can be a promising future research direction for efficient training. 

%% file: supp.tex
\maketitlesupplementary

\section{Implementation Details}

Table \ref{tab:hyparams} shows the detailed parameters used in the benchmarks (Main Paper Table 1 and 2). The rows above the horizontal line are the parameters that are also used for the existing methods, and the rest are the ones for our proposed method.
Gradient accumulation is used for ImageNet-Subset due to GPU memory constraint and the effective batchsize is displayed. 
At the initial-task stage, the model is trained with SGD as an optimizer with learning rate of 0.1 and weight decay of 5e-4 for 200 epochs, and at the following incremental stages with learning rate of 0.01 and weight decay of 2e-4 for 100 epochs (60 epochs for ImageNet-Subset). A cosine decay schedule is adopted and batch size is set to 64. For warm-start settings, the learning rate in the incremental stages is set to 0.001. All experiments are repeated three times and averaged, varying both the class-shuffling seed (1993 \cite{deepclsurvey}, 2993, 3993) and the randomness seed (0, 1000, 2000).

To avoid over-fitting to the test data, the shrinkage parameters $\gamma_1$ and $\gamma_2$ in Main Paper eq. 9 are determined by validation protocol, splitting the validation dataset ($N$=50 per class) from the train dataset for each setting (including ablation study). The candidates for $\gamma$s are discrete and the search space is (1, 3, 8, 16, 24, 32, 40, 48, 56, 64, 72, 80, 88, 96, 104, 112, 120). Table \ref{tab:gammas} shows the tuned $\gamma_1$ and $\gamma_2$ for FeCAM, AdaGauss and our APR.

For AdaGauss \cite{2024taskrecency}, we re-formulated the anti-collapse loss. The original loss $L_{AC}$ involves Cholesky decomposition of covariance matrix calculated from training minibatch. Cholesky decomposition fails when the target covariance is not  positive-difinite, especially at early training stage. We alternatively adopt eigenvalues of covariance as loss target instead of diagonal elements of Cholesky decomposed matrix. Our implementation is numerically stable and yields empirically comparative results with the original  $L_{AC}$. 
The output feature dimension is set to 64 following the original setting.

\begin{table}
    \begin{tabular}{c|cc}
         \toprule
         Augmentation & Parameter \\
        \hline
         RandomCrop & x, y, width, height  \\
         RandomResizedCrop & x, y, width, height \\
         RandomHolizontalFlip & do or not \\
         ColorJitter & function indices \\
         ColorJitter & magnitude values \\
         CIFAR10Policy & policy index \\
         ImageNetPolicy & policy index \\
         Autoaug policy 1 & do or not \\
         Autoaug policy 2 & do or not \\
         Autoaug ShearX & magnitude value\\
         Autoaug ShearY & magnitude value\\
         Autoaug TranslateX & magnitude value \\
         Autoaug TranslateY & magnitude value \\
         Autoaug Rotate & function index value \\
         Autoaug Color & magnitude \\
         Autoaug Contrast & magnitude value \\
         Autoaug Sharpness & magnitude value \\
         Autoaug Brightness& magnitude value \\
         \bottomrule
    \end{tabular}
    \caption{Augmentation random parameter list. The data type for "do or not" is \texttt{bool} and \texttt{int64} otherwise.}
    \label{tab:aug}
\end{table}

\section{Data Augmentation Parameters}
The augmentation parameter list is shown in Table \ref{tab:aug}. The random parameter values used in candidate selection are recorded and applied during new task training in a deterministic manner. 
During the candidate selection, AutoAugment \cite{cubuk2018autoaugment} policies for CIFAR10 and ImageNet are applied for CIFAR100 and ImageNet-Subset respectively, following random crop and horizontal flipping. In AutoAugment, two policies are randomly selected from 25 policies, each of which consists of two successive transforms. The magnitude and occurrence probability of each transform are fixed as the default values of AutoAugment. The augmentation pipelines are as follows:

\begin{itemize}
    \item CIFAR100: RandomCrop, RandomHolizontalFlip, ColorJitter, CIFAR10Policy.
    \item TinyImageNet: RandomCrop, RandomHolizontalFlip.
    \item ImageNet-Subset: RandomResizedCrop, RandomHolizontalFlip, ImageNetPolicy.
\end{itemize}

\begin{table*}
\centering
\resizebox{0.8\textwidth}{!}{

\begin{tabular}{cccccccccc}
\toprule

 & \multicolumn{4}{c}{CIFAR100}&\multicolumn{4}{c}{TinyImageNet}& ImageNetSubset \\
Parameter & \multicolumn{2}{c}{cold} & \multicolumn{2}{c}{warm} & \multicolumn{2}{c}{cold} & \multicolumn{2}{c}{warm} & cold  \\
\midrule
Number of tasks ($T$) & 5& 10& 6& 11& 5& 10& 6& 11& 10\\
Number of initial classes & 20& 10& 50 & 50 & 40& 20& 100 & 100 & 10\\
Number of incremental classes & 20& 10& 10& 5& 40& 20& 20& 10& 10\\
1st conv of extractor& \multicolumn{4}{c}{stride=2}& 
\multicolumn{4}{c}{stride=2}& stride=1 \\
Gradient accumulation& \multicolumn{4}{c}{1} & \multicolumn{4}{c}{1} & 2 \\
Number of epochs $T>1$ & \multicolumn{4}{c}{100} & \multicolumn{4}{c}{100} & 60 \\
KD loss weight $\lambda_{kd}$ & \multicolumn{4}{c}{10} & \multicolumn{4}{c}{10} & 10 \\
CE loss temperature & \multicolumn{4}{c}{1} & \multicolumn{4}{c}{1} & 1\\
Learning rate ($t=0$) & \multicolumn{4}{c}{0.1} & \multicolumn{4}{c}{0.1} & 0.1 \\
Learning rate ($t>0$) & \multicolumn{2}{c}{0.01} & \multicolumn{2}{c}{0.001} & \multicolumn{2}{c}{0.01} & \multicolumn{2}{c}{0.001} & 0.01 \\
Weight decay ($t=0$) & \multicolumn{4}{c}{5e-4} & \multicolumn{4}{c}{5e-4} & 5e-4 \\
Weight decay ($t>0$) & \multicolumn{4}{c}{2e-4} & \multicolumn{4}{c}{2e-4} & 2e-4 \\
Batchsize $B_{t}$ ($t=0$) & \multicolumn{4}{c}{64} & \multicolumn{4}{c}{64} & 64 \\
\hline
Batchsize $B_{t}$ ($t>0$) & \multicolumn{4}{c}{32} & \multicolumn{4}{c}{64} & 64 \\
Batchsize $B_{APR}$ ($t>0$) & \multicolumn{4}{c}{64} & \multicolumn{4}{c}{64} & 16\\
APR magnitude $\alpha$ & \multicolumn{4}{c}{64} & \multicolumn{4}{c}{64} &  64\\
APR iterations& \multicolumn{4}{c}{4} & \multicolumn{4}{c}{6} & 2 \\
APR number of candidates & \multicolumn{4}{c}{200} & \multicolumn{4}{c}{200} & 200 \\
ADC magnitude& \multicolumn{4}{c}{6.32} & \multicolumn{4}{c}{6.32} &3.16\\
ADC iterations& \multicolumn{4}{c}{9} & \multicolumn{4}{c}{6} & 3\\
ADC batchsize & \multicolumn{4}{c}{64} & \multicolumn{4}{c}{64} & 16 \\
ADC number of candidates & \multicolumn{4}{c}{1000} & \multicolumn{4}{c}{1000} & 1000\\
Transfer $W$ learning rate & \multicolumn{4}{c}{1e-4} & \multicolumn{4}{c}{1e-4} & 1e-4 \\
Transfer $W$ epochs & \multicolumn{4}{c}{64} & \multicolumn{4}{c}{64} & 64 \\
\bottomrule
\end{tabular}
}
\caption{Hyperparameter settings. The rows below the horizontal line are the settings regarding our method.}
\label{tab:hyparams}
\end{table*}

\begin{table*}
\centering
\resizebox{0.8\textwidth}{!}{
\begin{tabular}{cccccccccc}
\toprule
 & \multicolumn{4}{c}{CIFAR100}&\multicolumn{4}{c}{TinyImageNet}& ImageNetSubset \\
Parameter & \multicolumn{2}{c}{cold} & \multicolumn{2}{c}{warm} & \multicolumn{2}{c}{cold} & \multicolumn{2}{c}{warm} & cold  \\
\midrule
Number of tasks ($T$) & 5& 10& 6& 11& 5& 10& 6& 11& 10\\
Number of initial classes & 20& 10& 50 & 50 & 40& 20& 100 & 100 & 10\\
Number of incremental classes & 20& 10& 10& 5& 40& 20& 20& 10& 10\\
\hline
FeCAM \cite{goswami2024fecam} & 3& 1& 3& 3& 3& 1& 3& 3& 1\\
AdaGauss \cite{2024taskrecency} & 3& 3& 1& 3& 8 & 8& 3& 8& 8\\
APR (ours) & 16& 24& 40& 32 & 64& 40& 72 & 72 & 40\\
\bottomrule
\end{tabular}
}
\caption{Covariance shrinkage parameters $\gamma_1$ ($=\gamma_2$) tuned by the validation protocol, where the validation dataset ($N$=50 per class) is split from the train dataset.}
\label{tab:gammas}
\end{table*}

\section{Candidate Selection}\label{candidate}

In candidate sampling before each incremental task, $k$ data indices and augmentation policy parameters are sampled from the smallest feature-prototype distance. For example, 200 image indices are sampled from 5,000 new-task images for each class. Some images are assigned to multiple prototypes, especially at later stages. 
In this section we limit the number of assignments per new-task image and compare it with the main result setting where no limit is applied. 

To achieve this, we firstly calculate $(N, M)$ distance matrix $d_{mn}$ where $N$ is the number of new-task images and $M$ is the number of classes (prototypes). We sort all the values of $d_{mn}$ and assign the new-task image and the prototype from the minimum-distance pair. If the number of assigned classes for the image reaches $N_{cap}$ (e.g. 4) or the number of assigned images for the prototype reaches $k$ (e.g. 200), the distance value is skipped. The assignment is done when the numbers of assigned images for all the prototypes reach $k$.
The no-limit setting is considered as the spacial case of the above assignment where $N_{cap}=\infty$. 

Table \ref{tab:duplicate_cap} shows the comparison between $N_{cap}=4$ and $\infty$. We choose $N_{cap}=4$ because it is the minimum requirement for CIFAR100 setting at the final incremental stage : 200 image indices $\times$ 90 classes from 5,000 new-task images. 
As a result, there are no significant differences in performance, suggesting that diversity of sampled images does not affect the results because the pseudo-replay samples are diversified by adversarial attack toward various class prototypes (with noise).   
\begin{table}
\centering
\resizebox{0.68\linewidth}{!}{
    \begin{tabular}{c|cccc}
        \hline
         &  \multicolumn{2}{c}{NCM}& \multicolumn{2}{c}{Mahalanobis}\\ 
        $N_{cap}$ & $A_{inc}$ & $A_{last}$ & $A_{inc}$ & $A_{last}$ \\ 
        \hline
        4 & 68.95 & 55.95 & 69.92 & 57.65 \\ 
        $\infty$ & \textbf{68.93} & \textbf{56.23} & \textbf{69.96} & \textbf{57.94} \\ 
        \hline
    \end{tabular}
    }
    \vspace{-0.5em}
    \caption{Effect of candidate sampling limit per new-task image for CIFAR100 $T=10$ setting. The baseline results from Main Paper Tab. 1 are highlighted in bold.}
    \label{tab:duplicate_cap}
\end{table}

\section{Covariance Decomposition}\label{sec:decomposition}
For the covariance decomposition experiment in Sec. 4.5 of Main Paper, we use the ablation study setting without prototype augmentation noise ($r=0$) as the baseline setting, because the noise magnitude $r$ depends on covariance matrices (eq. 4 in Main Paper). To avoid irrelevant fluctuation, the saved network weights checkpoint at each task of the baseline setting is used, and only covariance decomposition setting is changed (with and without decomposition, different $k$ values).
The matrices $U, S, V$ are re-composed into a full covariance matrix only during covariance calibration and Mahalanobis distance calculation at test time. Covariance shrinkage parameters $\gamma_1$ and $\gamma_2$ are determined for each $k$ setting similar to the benchmark settings.

\section{Storage Calculation}\label{sec:storagedetails}
Required storage in Table 7 (Main Paper) is calculated with the following assumptions:
\begin{itemize}
    \item Benchmark: ImageNet-Subset, $T=10$, $t=9$ (final stage)
    \item Prototypes: 90 classes, 512-dim (64-dim for AdaGauss)
    \item Covariances: 90 classes, 512-dim (64-dim for AdaGauss)
    \item Sample inds: 200 candidates $\times$ 90 classes, in \texttt{int64}
    \item Aug. params: (10 \texttt{int64} and 3 \texttt{bool} params) $\times$ 13k images
    \item Actual image data (for comparison, Sec. 4.5): 200 samples $\times$ 90 classes, each (3, 224, 224) in \texttt{float32}
\end{itemize}

\section{Training time}

We compare detailed wall-clock measurements among methods in Table \ref{tab:wallclock}. Due to online adversarial attack, APR's training time at incremental tasks is larger than the rest.

\begin{table}[h!]
\centering
    \resizebox{0.96\linewidth}{!}{
    \begin{tabular}{c|cccc}
        \hline
        Component & FeCAM & ADC & AdaGauss & APR  \\
        \hline
        Init task $t=0$ & 2.8 & 2.8 & 2.8 & 2.8 \\
        Inc. task $t=9$ & 0.0 & 1.1 & 1.1 & 2.9   \\
        Calibration (90 cls) & 0.00 & 0.10 & 0.84 & 0.36 \\
        \hline
        Total benchmark & 2.9 & 13.5 & 20.1 & 31.0 \\
        \hline
    \end{tabular}
}
\caption{Comparison of training time (hour) at $t=0$, $t=9$ and calibration on ImageNet-Subset $T=10$, averaged across three-seed runs.}
\label{tab:wallclock}
\end{table}

\section{Visualization}

\begin{figure}
  \centering

  \begin{subfigure}[b]{0.99\linewidth}
    \includegraphics[width=\linewidth]{figs/tsne_attack.pdf}
    \vspace{-2em}
    \caption{$t=1$}
  \end{subfigure}\

  \begin{subfigure}[b]{0.99\linewidth}
    \includegraphics[width=\linewidth]{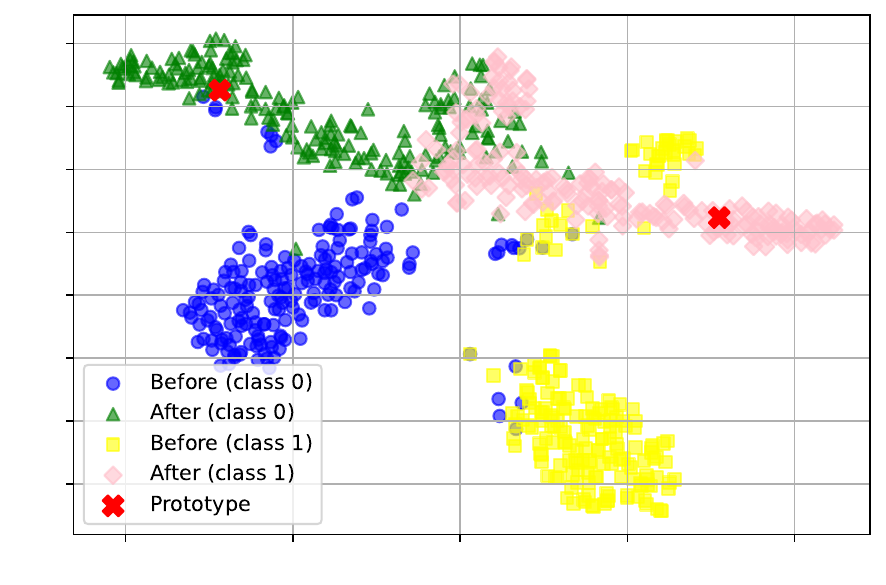}
    \vspace{-2em}
    \caption{$t=5$}
  \end{subfigure}

  \begin{subfigure}[b]{0.99\linewidth}
    \includegraphics[width=\linewidth]{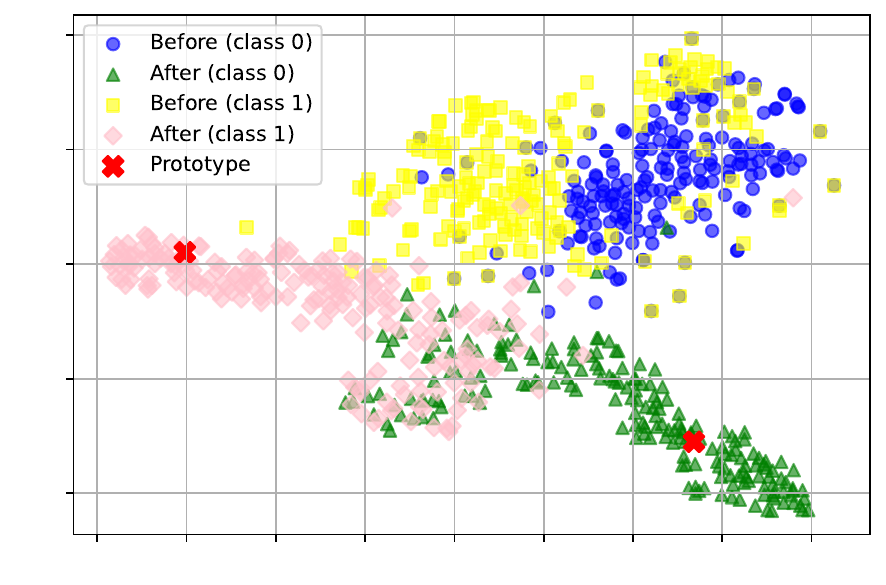}
    \vspace{-2em}
    \caption{$t=9$}
  \end{subfigure}

  \vspace{-1.0em}
  \caption{t-SNE feature distribution analysis during $t=1, 5, 9$ on CIFAR100. The perturbation effect persists both in the early and late incremental stages. }
  \label{fig:tsne_3stages}
\end{figure}

Fig. \ref{fig:tsne_3stages} visualizes the t-SNE plot of feature distributions before and after perturbation at $t=1,5,9$ on CIFAR100. The perturbation effect persists in the late incremental stage.

\section{Reproducibility checklist}

We will make our code available when accepted. Our repository guarantees reproducibility including seed control and recording function. The following the reproducibility checklist based on \cite{pineau2020reproducibility}.

\begin{itemize}
    \item Specification of dependencies: \textbf{Yes. We provide docker environment for reproducible dependencies.}
    \item Training code: \textbf{Yes.}
    \item Evaluation code: Yes. 
    \item Config files: \textbf{Yes. We provide hierarchical configs.}
    \item (Pre-)trained model(s): \textbf{Yes.} 
    \item README file includes table of results accompanied by precise command to run: \textbf{Yes.}
    \item Seed control: \textbf{Yes, deterministic training with a class-shuffling seed and a randomnesses seed is possible.}
    \item Exact augment parameter recording format: \textbf{Yes. We record all the config parameters, precise command, wall time, git SHA and all the necessary metrics (at every incremental stage) in csv format.}
\end{itemize}